\documentclass[letterpaper, 10 pt, conference]{ieeeconf}  

\IEEEoverridecommandlockouts                              

\usepackage{amsmath,amsfonts}
\usepackage{algorithm}
\usepackage{algpseudocode}

\usepackage{array}
\usepackage[caption=false,font=normalsize,labelfont=sf,textfont=sf]{subfig}
\usepackage{textcomp}
\usepackage{stfloats}
\usepackage{url}
\usepackage{verbatim}
\usepackage{graphicx}
\usepackage{xcolor}
\usepackage{makecell}
\usepackage{cite}
\usepackage{colortbl}

\definecolor{cvprblue}{rgb}{0.21,0.49,0.74}
\definecolor{ForestGreen}{RGB}{34,139,34}
\usepackage[
        breaklinks,
        colorlinks,
        citecolor=blue,
        urlcolor=black,
        linkcolor=blue,
        ]{hyperref}
\usepackage{multirow}

\title{\LARGE \bf
Directed-CP: Directed Collaborative Perception for Connected and Autonomous Vehicles via Proactive Attention
}

\author{Yihang Tao\textsuperscript{\rm 1}, Senkang Hu\textsuperscript{\rm 1}, Zhengru Fang\textsuperscript{\rm 1}, and Yuguang Fang\textsuperscript{\rm 1*}\thanks{*Corresponding author. The research work described in this paper was conducted in the JC STEM Lab of Smart City funded by The Hong Kong Jockey Club Charities Trust under Contract 2023-0108. The work was also supported in part by the Hong Kong SAR Government under the Global STEM.}
 \thanks{$^{1}$Yihang Tao, Senkang Hu, Zhengru Fang and Yuguang Fang are with Department of Computer Science,
         City University of Hong Kong, Kowloon, Hong Kong.
         (Email: {\tt\footnotesize \{yihangtao2-c, senkang.forest, zhefang4-c\}@my.cityu.edu.hk, my.Fang@cityu.edu.hk})}%
}

\begin{document}

\maketitle
\thispagestyle{empty}
\pagestyle{empty}

\begin{abstract}


Collaborative perception (CP) leverages visual data from connected and autonomous vehicles (CAV) to expand an ego vehicle's field of view (FoV). Despite recent progress, current CP methods do expand the ego vehicle's 360-degree perceptual range almost equally, but faces two key challenges. Firstly, in areas with uneven traffic distribution, focusing on directions with little traffic offers limited benefits. Secondly, under limited communication budgets, allocating excessive bandwidth to less critical directions lowers the perception accuracy in more vital areas.
To address these issues, we propose Directed-CP, a proactive and direction-aware CP system aiming at improving CP in specific directions. Our key idea is to enable an ego vehicle to proactively signal its interested directions and readjust its attention to enhance local directional CP performance.
To achieve this, we first propose an RSU-aided direction masking mechanism that assists an ego vehicle in identifying vital directions. Additionally, we design a direction-aware selective attention module to wisely aggregate pertinent features based on ego vehicle's directional priorities, communication budget, and the positional data of CAVs. Moreover, we introduce a direction-weighted detection loss (DWLoss) to capture the divergence between directional CP outcomes and the ground truth, facilitating effective model training.
Extensive experiments on the V2X-Sim 2.0 dataset demonstrate that our approach achieves 19.8\% higher local perception accuracy in interested directions and 2.5\% higher overall perception accuracy than the state-of-the-art methods in collaborative 3D object detection tasks. Codes are available at \texttt{https://github.com/yihangtao/Directed-CP.git}.

\end{abstract}

\section{Introduction}

Collaborative perception (CP)\cite{fang2024pacp, hu2024fullscenedomaingeneralizationmultiagent, zhang2024smartcoopervehicularcollaborativeperception} has emerged as a promising approach to expand the perceptual range of individual vehicles by integrating visual data from multiple connected and autonomous vehicles (CAVs). To effectively monitor road traffic, each CAV is equipped with an array of LiDARs or cameras that capture environmental data from various angles. This information is subsequently synthesized into a bird's eye view (BEV) map, offering a comprehensive representation of a vehicle's surroundings\cite{10160968}. Nonetheless, relying solely on a single BEV-aided perception system is often insufficient for overcoming blind spots caused by road obstacles or other CAVs. To address this shortcoming, CP has been adapted to allow multiple CAVs to share their local BEV features, thereby enhancing the accuracy and coverage of BEV predictions.

\begin{figure}[t]
    \centering
    \includegraphics[width=0.9\linewidth]{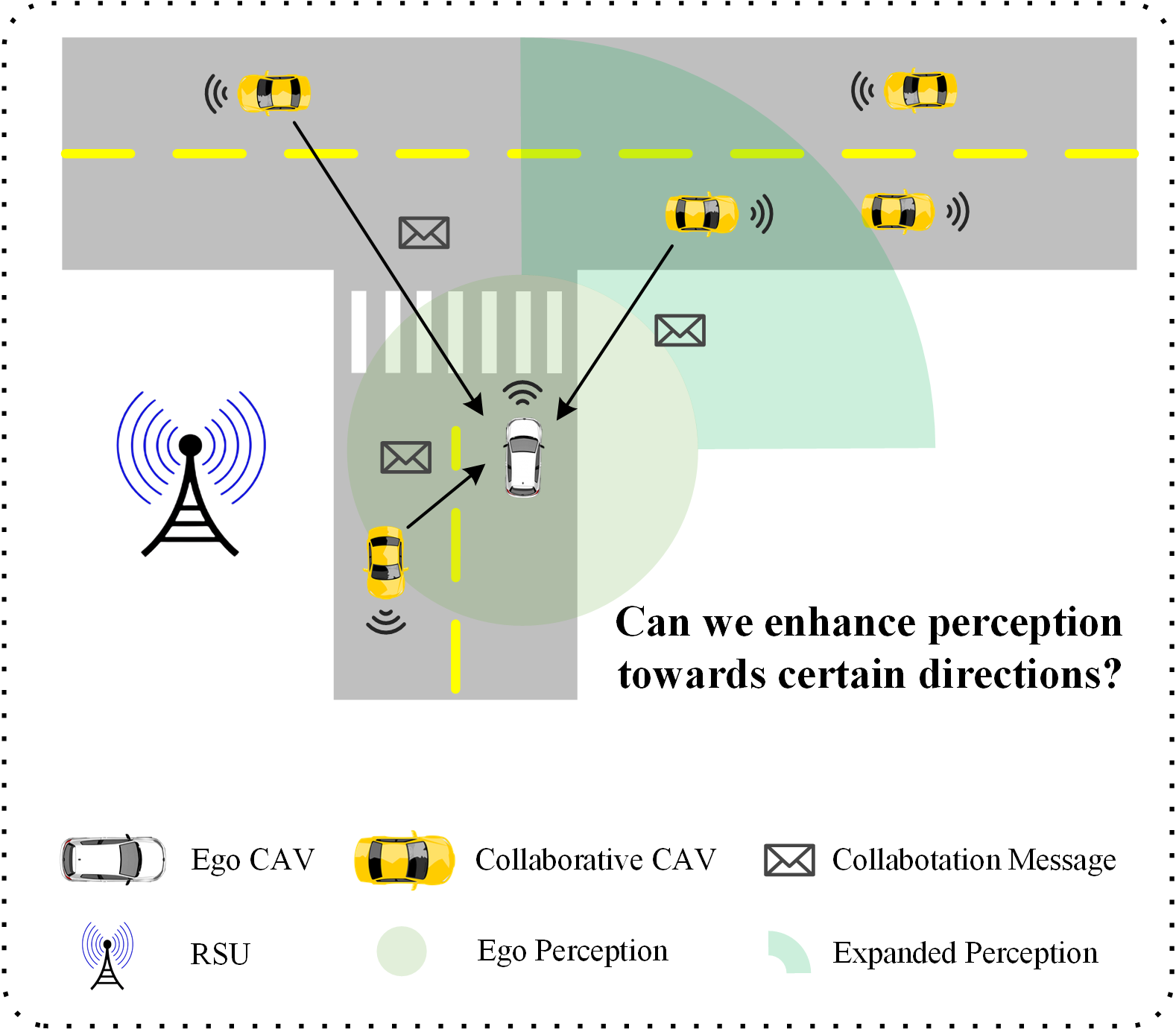}
    \caption{\textbf{Overview of directed CP framework.} Under limited communication budget, ego CAV enhances perception in complex traffic directions while maintaining basic awareness in low-traffic areas.}
    \label{fig: scenario}
    \vspace{-3mm} 
\end{figure}

Currently, most existing studies \cite{Liu_2020_CVPR, li2021learning} focus on optimizing 360-degree omnidirectional CP performance, aiming to extend an ego CAV's scope in every direction almost equally. However, this overlooks the non-uniform traffic density across different directions and varying interests of an ego CAV in specific directions. For instance, as illustrated in Fig. \ref{fig: scenario}, when an ego CAV is making a right turn at an intersection, it may encounter minimal traffic to its rear and left front, whereas the traffic is significantly more complex to its right front. In such scenarios, the ego CAV would benefit from a targeted perception enhancement towards its right front while maintaining basic (e.g., single-vehicle) perception for other directions. Existing methods aim to uniformly enhance perception across all directions, lacking the flexibility for an ego CAV to proactively adjust its view-level priority.

In addition, communication overhead is critical factor that must be carefully considered when designing a CP system\cite{9197364, 10.1007/978-3-030-58536-5_36,hu2024fullscenedomaingeneralizationmultiagent,huCollaborativePerceptionConnected2024, hu2022wherecomm, huAgentsCoDriverLargeLanguage2024, hu2024agentscomergelargelanguagemodel,fang2022age}.  With constraints such as a limited communication budget and a maximum allowable delay, engaging all collaborators and fully utilizing their captured views for enhancing perception across 360 degrees can significantly burden both communication and computational resources. This is particularly severe when the number of collaborators and the frame rate (measured in frames per second, FPS) are high. Indeed, reallocating communication resources from less critical directions to enhance perception in more important areas is not only strategically advantageous but also enhances CP efficiency in terms of communication and computation.

Motivated by the above observation, we propose Directed-CP, which enables an ego CAV to proactively specify its interested directions and intelligently optimize perception performance toward these directions under constraints. To achieve this, we propose to deploy several roadside units (RSUs) to monitor the traffic distribution around an ego CAV. These RSUs provide critical data that assists the ego vehicle in determining its interested directions. Additionally, we have developed a direction-aware attention module that inputs the ego CAV's preferred directions, communication budget, and the positional information of other CAVs, thereby generating sparse query maps that can intelligently select the most relevant information from nearby CAVs for fusion to enhance CP performance toward the selected directions. Moreover, we define a direction-weighted detection loss (DWLoss) to measure the directional perception discrepancy between prediction and ground truth. To the best of our knowledge, this is the first work designed to optimize CP based on local directional priorities. Our contributions are summarized as follows.
\begin{itemize}
\item We propose a flexible CP framework named Directed-CP, which enhances perception performance towards specific directions under communication budget, tailored to the proactive interests of an ego CAV.
\item We design a direction-aware selective attention module to incorporate an RSU-aided direction masking mechanism, and adaptively select relevant feature data from multi-vehicle to boost local-directional perception. Additionally, we design a direction-weighted detection loss (DWLoss) to measure the directed perception discrepancy between the outputs and the ground truth.
\item We conduct extensive experiments on collaborative 3D detection tasks and demonstrate that our method realizes the proactive directed CP enhancement, achieving 2.5\% higher overall perception accuracy and 19.8\% higher local perception accuracy in the interested directions than the state-of-the-art method.
\end{itemize}
\section{Related Works}

\begin{table}[t]
    \caption{Comparison of related works.}
    \label{tab:related_works}
    \renewcommand{\arraystretch}{1.15}
    \setlength{\tabcolsep}{3pt}
    \resizebox{1\linewidth}{!}{
    \begin{tabular}{l|ccc}
        \hline
        \textbf{Method} & \textbf{Message} & \textbf{Fusion} & \textbf{Perception Gain} \\
        \hline\hline
        Who2com~\cite{liu2020who2comcollaborativeperceptionlearnable} & Full & Average & Omnidirectional \\
        V2VNet~\cite{10.1007/978-3-030-58536-5_36} & Full & Average & Omnidirectional \\
        PACP~\cite{fang2024pacp} & Full & Priority-based & Omnidirectional \\
        When2com~\cite{Liu_2020_CVPR} & Full & Agent attention & Omnidirectional \\
        V2X-ViT~\cite{10.1007/978-3-031-19842-7_7} & Full & Self-attention & Omnidirectional \\
        Where2com~\cite{hu2022wherecomm} & Sparse & Location attention & Omnidirectional \\
        \rowcolor{gray!5}
        \textbf{Directed-CP (Ours)} & Sparse & Proactive attention & \textbf{Directed} \\
        \hline
    \end{tabular}
    }
\end{table}

\subsection{Collaborative perception}

Collaborative perception extends vehicles' sensing capabilities beyond single-vehicle's limits through intermediate-stage fusion strategies \cite{9676458, 10160546, 10160367, 10161460, 10160871}. While this enables feature exchange among CAVs, increasing feature dimensions and the number of collaborator demand efficient bandwidth management.
Who2com \cite{liu2020who2comcollaborativeperceptionlearnable} employs a multi-stage handshake mechanism to compress information via matching scores, while V2VNet \cite{10.1007/978-3-030-58536-5_36} uses graph neural networks to aggregate information from nearby CAVs. These methods, however, neglect the varying importance of individual CAVs. PACP \cite{fang2024pacp} addresses this limitation with a BEV-match mechanism for CAV prioritization, but it only considers agent-level priorities while ignoring view-specific importance. Furthermore, PACP focuses on omnidirectional perception but lacks the flexibility for directional perception adjustment, which is the focus of this paper.

\subsection{Attention-based LiDAR perception}

Recent advancements in LiDAR-based CP have integrated attention mechanisms to boost performance and reduce communication overhead.
When2com \cite{Liu_2020_CVPR} employs scaled general attention to assess correlations among different agents, reducing transmission redundancy. V2X-ViT \cite{10.1007/978-3-031-19842-7_7} introduces the heterogeneous multi-agent attention for fusing messages across diverse agents. However, these methods require initial transmissions of full feature maps, which consumes substantial bandwidth.
More recently, Where2comm \cite{hu2022wherecomm} advances the field by utilizing sparse feature maps with location-specific and confidence-aware attention, optimizing data exchange and processing efficiency by focusing on the most relevant features. Despite its progress, Where2comm lacks the flexibility for an ego vehicle to adjust its perceptual focus based on immediate environmental demands and may not be effective under communication constraints.
As outlined in Table \ref{tab:related_works}, our proposed Directed-CP contrasts by providing a flexible and directed perception enhancement tailored to an ego vehicle's proactive needs under limited communication constraints. This targeted approach improves data relevance and efficiency, aligning closely with real-time needs under dynamic vehicular settings.
\section{Methodology}

\begin{figure*}[t]
    \centering
    \includegraphics[width=0.9\linewidth]{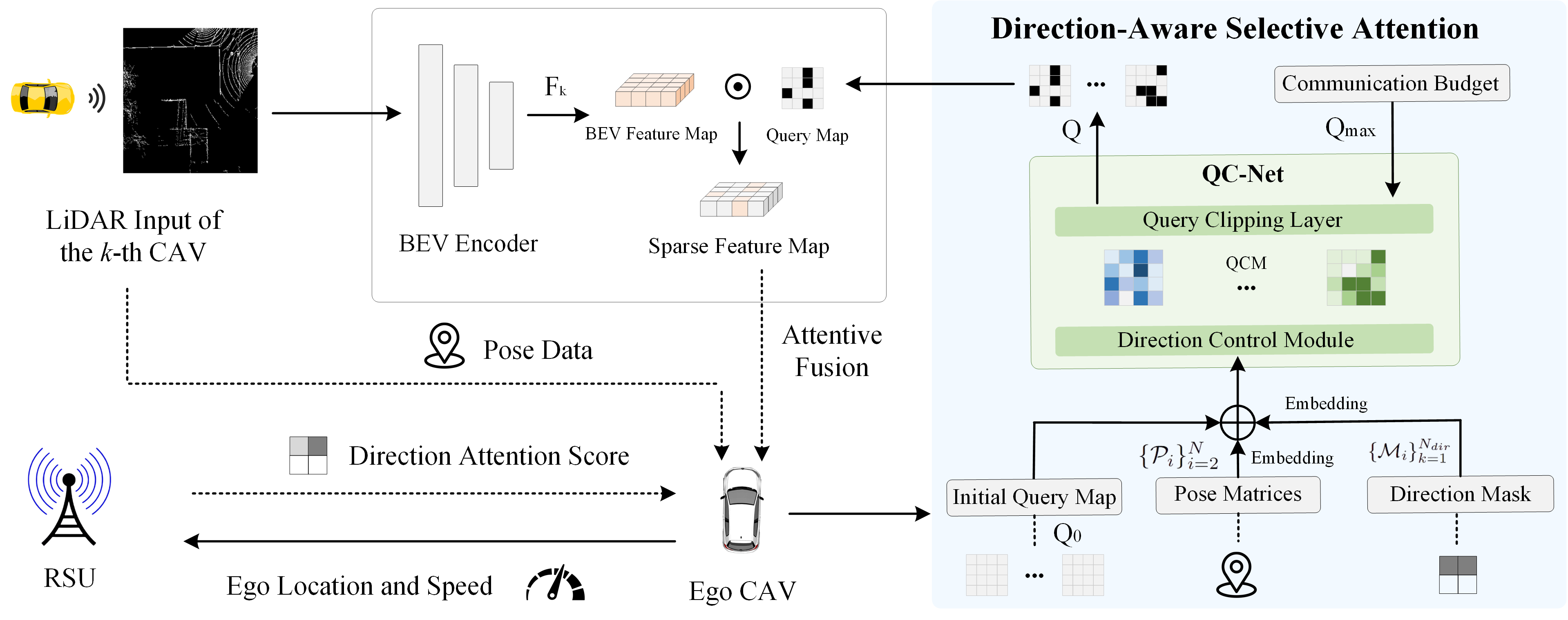}
    \caption{\textbf{Problem overview.} Ego CAV combines RSU's DAS with its interests to create direction mask. The mask, along with initial query map, neighboring CAVs' poses, and communication budget, are input to QC-Net. QC-Net contains: (i) \textbf{Direction Control Module,} generating direction-prioritized query confidence maps (QCMs), and (ii) \textbf{Query Clipping Layer,} selecting top $Q_{max} \times H \times W$ queries based on QCMs under budget constraints.}
    \label{fig: method}
\end{figure*}

Fig.~\ref{fig: method} illustrates our method's architecture. RSUs deployed along the roadway monitor traffic from elevated positions, providing broader views than individual vehicles. An ego CAV periodically exchanges its status with nearby RSUs to receive direction attention scores (DAS). Based on DAS and its interests, the ego CAV masks non-essential directions during collaborative perception. Then, guided by prioritized directions, communication budget, and neighboring CAVs' poses, it selects optimal feature map queries to maximize directed perception performance. The following subsections detail these components.

\subsection{RSU-aided direction masking}

In this paper, we leverage RSUs to help ego CAVs identify important directions. The 360-degree space around an ego CAV is divided into $N_{dir}$ local directions. Based on ego CAV's location and speed, the corresponding RSU projects it into its captured 2D view and calculates Direction Attention Scores (DAS) for each direction.
For DAS calculation, we primarily use vehicle density as an indicator, as areas with higher vehicle densities typically require more attention. The DAS from RSU is represented as $\{\mathcal{S}_{r}^{i}\}_{k=1}^{N_{dir}} = \{N_{vec}^{i}\}_{k=1}^{N_{dir}}$, where $N_{vec}^{i}$ denotes the number of detected vehicles in the $i$-th direction. When RSU is unavailable, the system can alternatively use historical CP results to estimate traffic density, ensuring system robustness. This approach can be extended to incorporate additional factors such as vehicle speeds, accident history, and road conditions.
The ego CAV then combines RSU's DAS $\{\mathcal{S}_{r}^{i}\}_{i=1}^{N_{dir}}$ with its own interest weights $\{\mathcal{I}_{e}^{i}\}_{i=1}^{N_{dir}}$ to calculate the final direction mask. These interest weights can be flexibly adjusted according to the ego CAV's proactive needs. In cases where the weights are uniformly assigned, the direction importance relies entirely on RSU's DAS. The final direction mask $\{\mathcal{M}_{i}\}_{i=1}^{N_{dir}}$ is calculated as follows:
\begin{equation}
    \mathcal{M}_{i} = \max\left\{H\left(\frac{\mathcal{S}_{r}^{i} \mathcal{I}_{e}^{i}}{\sum_{j=1}^{N_{dir}} \mathcal{S}_{r}^{j} \mathcal{I}_{e}^{j}}-\sigma_1\right), H(\mathcal{S}_{r}^{i} \mathcal{I}_{e}^{i} - \sigma_2)\right\},
\end{equation}
where Heaviside step function $H(\cdot)$ equals 1 for positive inputs and 0 otherwise. The threshold $\sigma_1$ determines the relative importance of each direction, while $\sigma_2$ serves as an absolute threshold to identify complex traffic scenarios even in relatively less important directions.

Our RSU-aided direction masking mechanism offers several key advantages. First, the communication between an ego CAV and an RSU only involves basic information (location, speed, and DAS), ensuring minimal bandwidth usage and real-time performance. Second, the interest weight matrix provides ego CAVs with full autonomy of direction prioritization, allowing them to overwrite RSU suggestions when necessary. Third, the dual-threshold design ($\sigma_1$ and $\sigma_2$) enables both relative and absolute traffic complexity assessment, enhancing the system's adaptability to various scenarios.

\subsection{Direction-aware selective attention}

Consider $N$ CAVs in total in the scenario. Assume that the direction priority, the observation sets and perception supervision of the $i$-th CAV are represented as $\mathcal{M}_i$, $\mathcal{X}_{i}$ and $\mathcal{Y}_i$, respectively. {\color{black}Let $\xi_{\Phi}(B,N) = (\theta^*, \mathcal{T}^*)$ denote the optimal parameters and transmission strategy that maximize the directed perception performance.} The objective of our considered directed collaborative perception system is to achieve the maximized perception performance toward interested directions of all agents as a function of communication budget $B$ and the number of CAVs $N$, written as:
\begin{equation}
    \begin{aligned}
        \xi_{\Phi}(B,N) & = \arg\max_{\theta,\mathcal{T}} \sum_{i=1}^{N}g\left(\Phi_{\theta}\left(\mathcal{X}_{i}, \{\mathcal{T}_{i, k}\}_{k=1}^{N}\right), \mathcal{M}_i, \mathcal{Y}_i\right), \\
        & \mathrm{s.t.} \ \sum_{k=1}^{N} |\{\mathcal{T}_{i, k}\}_{k=1}^{N}| \leq B,
    \end{aligned}
\end{equation}
where $g(\cdot, \cdot)$ is the perception performance metric, $\Phi_{\theta}$ is the perception model with trainable parameter $\theta$, $\{\mathcal{T}_{i, k}\}_{k=1}^{N}$ are the messages transmitted from the $k$-th agent (each with $M$ features) to the $i$-th agent. Note that the case when $N=1$ indicates single-vehicle perception.

Upon receiving a 3D point cloud, the $i$-th CAV first converts the data into a BEV map. The BEV encoder $\Phi_{bev}$ processes this map to extract features, generating the feature map $\Phi_{bev}(\mathcal{X}_i) = \mathcal{F}_i \in \mathbb{R}^{H \times W \times D}$, where $H$, $W$, and $D$ represent height, width, and channel dimensions, respectively.
All agents project their perceptual data into a unified global coordinate system, facilitating seamless cross-agent collaboration without the need of complex coordinate transformations. The resulting feature map is fused with each other following direction-aware selective attention (DSA).
The core component of DSA is the query-control net (QC-Net) taking initial query map $\mathcal{Q}_{0} \in \mathbb{R}^{H \times W \times (N-1)}$, the embedding of nearby cooperative CAVs' pose matrices $\text{PE}(\{\mathcal{P}_i\}_{i=2}^{N}) \in \mathbb{R}^{H \times W \times (N-1)}$, the embedding of the ego CAV's direction mask $\text{DE}(\{\mathcal{M}_{k}\}_{k=1}^{N_{dir}}) \in \mathbb{R}^{H \times W \times (N-1)}$, and the communication budget as input, and generates proactive binary query maps $\{\mathcal{Q}_{k}\}_{k=2}^{N} \in \mathbb{R}^{H \times W \times (N-1)}$  (value 1 means activating transmitting data in the corresponding location of BEV feature map). The \textbf{communication budget} $Q_{max} \in [0,1]$ is defined as the ratio of the maximum number of activated queries to the size of the query map satisfying:
\begin{equation}
    \label{budget}
    Q_{max} \ge \sum_{k=2}^{N} \sum_{i=1}^{H} \sum_{j=1}^{W} \frac{\mathcal{Q}_{k}^{i,j}}{H \times W \times (N-1)},
\end{equation}
where $Q_{max}=1$ means allowing CAVs to transmit full feature map to the ego CAV. The QC-Net consists of a three-layer MLP. The direction control module first generates query confidence map (QCM) $\{\mathcal{C}_{k}\}_{k=2}^{N} \in \mathbb{R}^{H \times W \times (N-1)}$ for each CAV, while $\mathcal{C}_k^{i,j} \in [0,1]$ represents the priority of the $j$-th element of the $i$-th QCM for enhancing CP in the ego CAV's interested directions. Assume the direction control module is denoted with $\Phi_{dcl}(\cdot)$, QCM is calculated by:
\begin{equation}
    \{\mathcal{C}_{k}\}_{k=2}^{N} = \Phi_{dcl} \left( \mathcal{Q}_{0}, \text{PE}(\{\mathcal{P}_i\}_{i=2}^{N}), \text{DE}(\{\mathcal{M}_{k}\}_{k=1}^{N_{dir}}) \right).
\end{equation}
Given communication constraints, we introduce a query clipping layer to control the transmitted data during the collaboration.  In this layer, we rank $\mathcal{C}_k^{i,j}$ for each QCM, retaining only the top $Q_{max} \times H \times W$ values and setting others to zero, ensuring adherence to the predefined communication budget. The QC-Net finally produces sparse query maps $\{\mathcal{Q}_{k}\}_{k=2}^{N}$ as follows:
\begin{equation}
    \mathcal{Q}_k^{i,j} = 
    \begin{cases} 
    1, & \text{if } \mathcal{C}_k^{i,j} \in \text{TOP}_{Q_{max} \times H \times W}\left(\{\mathcal{C}_{k}\}_{k=2}^{N}\right), \\ 
    0, & \text{otherwise,}
    \end{cases}
\end{equation}
where $\text{TOP}_{k}(\cdot)$ reprsents the top $k$ elements of a set.

Collaborative CAVs receive these query maps and compute direction-aware sparse feature maps as $\mathcal{H}_{i} = \mathcal{Q}_{i} \odot \mathcal{F}_{i} \in \mathbb{R}^{H \times W \times D}$, where $\odot$ denotes the Hadamard product of two matrices. Subsequently, each ego CAV fuses features from multiple agents at each spatial location:
\begin{equation}
    \label{weights}
    W_{i,j}^{DSA} = \text{MAttn}\left(\mathcal{F}_{i}, \mathcal{H}_{i,j}, \mathcal{H}_{i,j}\right) \odot  \mathcal{C}_j,
\end{equation}
where $W_{i,j}^{DSA} \in  \mathbb{R}^{H \times W}$ is DSA weights assigned to the $j$-th agent by the $i$-th agent, $\text{MAttn}(\cdot)$ represents the multi-head attention at each spatial location. The fused feature map for the ego CAV $\mathcal{F}_{i}^{out} \in  \mathbb{R}^{H \times W \times D}$ is expressed as:
\begin{equation}
    \mathcal{F}_{i}^{out} = \text{FFN}\left(\sum_{j=1}^{N} W_{i,j}^{DSA} \odot  \mathcal{H}_{i,j}\right),
\end{equation}
where $\text{FFN}(\cdot)$ is the feed-forward network. 

\subsection{Direction-weighted detection loss}

Given the final fused feature map $\mathcal{F}_{i}^{out}$, the detection decoder $\Phi_{dec}(\cdot)$ generates class and regression outputs following \cite{hu2022wherecomm}. Each output location $\Phi_{dec}(\mathcal{F}_{i}^{out}) \in \mathbb{R}^{H \times W \times 7}$ corresponds to a rotated box described by a 7-tuple $(c,x,y,h,w,\cos\alpha, \sin\alpha)$, representing class confidence, position, size, and angle. {\color{black}The angle is parameterized using both sine and cosine components to ensure continuous optimization and avoid the periodicity issue at $\pm\pi$.}
To evaluate the discrepancy between the collaborative 3D detection results and the ground truth, the commonly used detection loss $\mathcal{L}_{det}$ \cite{zhou2019objectspoints} combines focal loss, object offset loss, and object size loss. However, this loss does not fully capture the importance of specific directions in our directed CP scenario. Therefore, we introduce a novel loss function, direction-weighted detection loss (DWLoss), to quantify the divergence in designated directions.
DWLoss is calculated by dividing the 3D detection results into $N_{dir}$ subsets and computing the detection loss $\{\mathcal{L}_{det}^{i}\}_{i=1}^{N_{dir}}$ for each subset with varying sum weights, represented as follows:
\begin{equation}
    \mathcal{L}_{DW} = \frac{\sum_{i=1}^{N_{dir}} \mathcal{L}_{det}^{i}*(\mathcal{M}_i+\sigma)}{\sum_{i=1}^{N_{dir}} \mathcal{M}_i + \sigma N_{dir}},
    \label{ldwn}
\end{equation}
where $\sigma$ is a constant weight-control factor. Eq. \ref{ldwn} ensures that lower weights are assigned to non-critical directions by weight factor $\sigma$, aiming to jointly optimize the CP performance in interested directions and the remaining directions. The choice of $\sigma$ is crucial: too high may obscure the importance of interested directions, while too low (an extreme case is 0) can neglect the accuracy in non-critical directions during training, potentially degrading perception more than single-vehicle perception. Ablation studies in Section IV will offer helpful guidance for determining an effective $\sigma$.

\section{experiments}

\begin{figure}[t]
    \centering
    \includegraphics[width=0.95\linewidth]{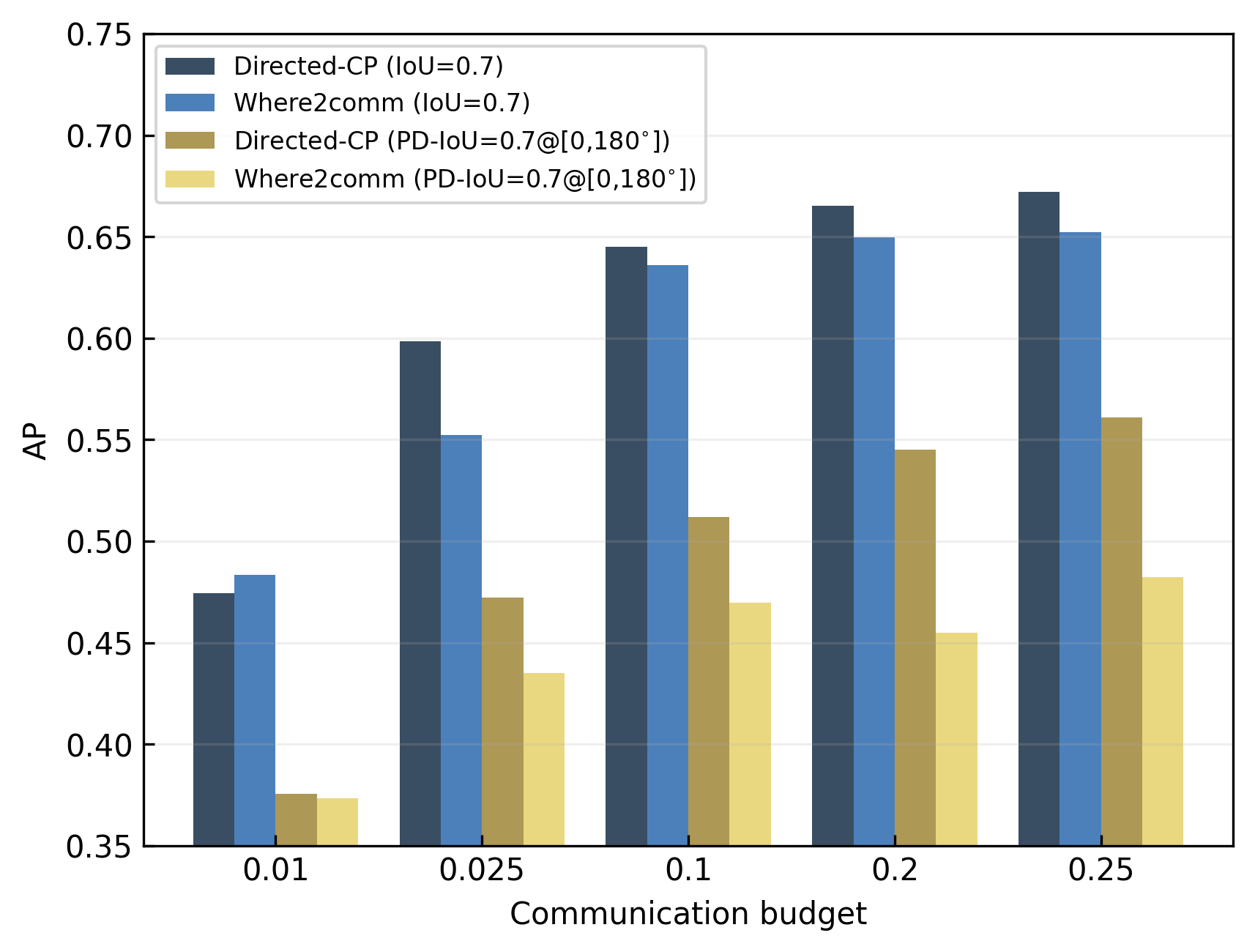}
    \caption{\textbf{AP of different CP methods} under various communication budgets.}
    \label{fig: experiment}
\end{figure}

\begin{figure*}[t]
    \centering
    \includegraphics[width=0.96\linewidth]{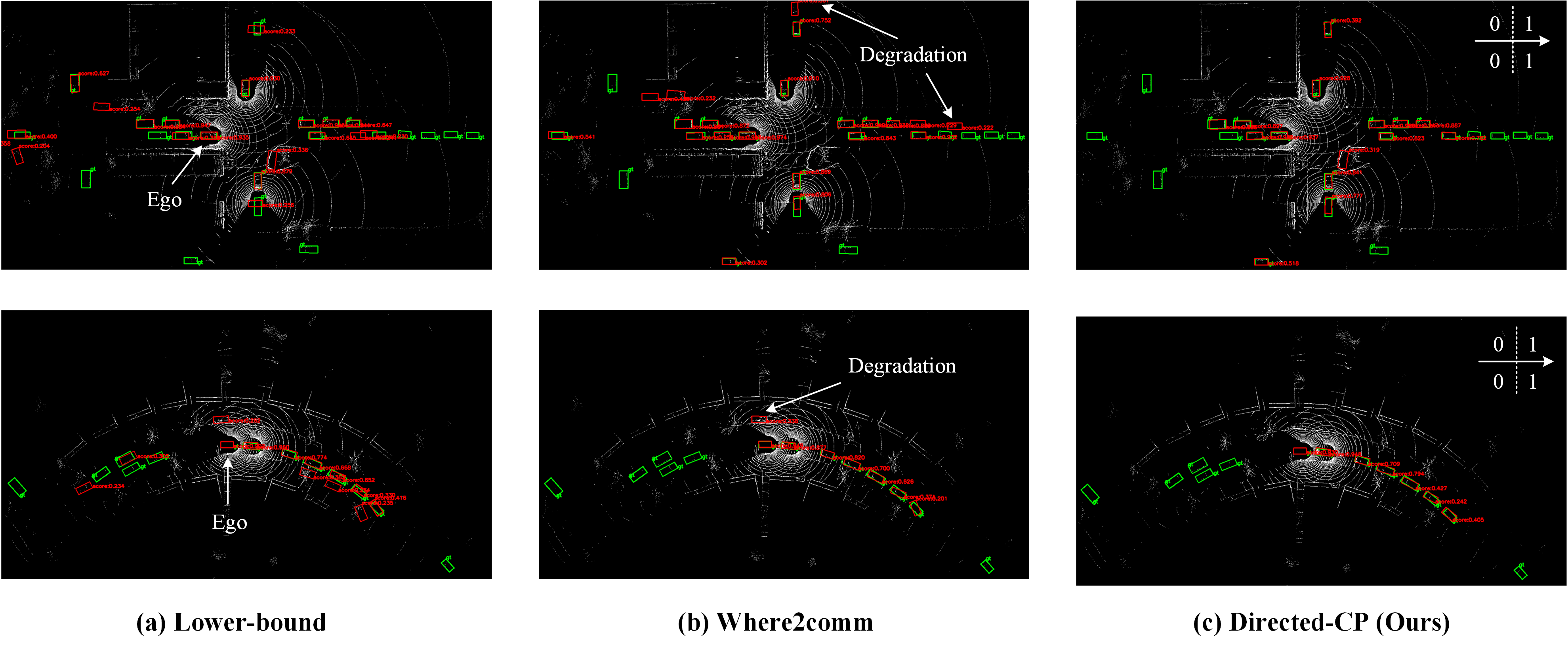}
    \caption{\textbf{Visualization comparison} on V2X-Sim 2.0 dataset. \textcolor{ForestGreen}{Green} and \textcolor{red}{red} boxes denote ground truth and predictions, respectively. While Where2comm improves global perception over lower-bound, it shows limitations in certain directions. Directed-CP enhances perception in ego's interested directions (marked as 1, with right arrow showing ego CAV's movement).}
    \label{fig: visualization}
    \vspace{-3mm} 
\end{figure*}

\subsection{Experimental setup}

\textbf{Dataset and baselines.}
Our experimental evaluations are conducted on the V2X-Sim 2.0 Dataset \cite{li2022v2xsimmultiagentcollaborativeperception}, an extensive simulated dataset generated using the CARLA simulator \cite{pmlr-v78-dosovitskiy17a}. This dataset comprises 10,000 frames of 3D LiDAR point clouds along with 501,000 annotated 3D bounding boxes. We configure the perception range to be $64m \times 64m$, and the 3D points are discretized into a BEV map of dimensions $(252, 100, 64)$.
We establish baseline comparisons, including When2com \cite{Liu_2020_CVPR}, V2VNet \cite{10.1007/978-3-030-58536-5_36}, and Where2comm \cite{hu2022wherecomm}. To make the perception gain clearer, we set the single-vehicle perception method as the lower-bound baseline. 

\textbf{Implementation details.}
We implement our Directed-CP using PyTorch. The direction control module features a fully connected layer with dimensions of $100\times252$ for both input and output, complemented by a sigmoid layer to match the BEV feature dimensions, incorporating pose matrices and direction masks at a spatial resolution of $(100, 252)$. Our detection module utilizes the LiDAR-based 3D object detection framework PointPillar \cite{Lang_2019_CVPR}. We set the training batch size at 6 and the maximum epochs at 60. The 360-degree space is divided into 4 directions: [0, 90$^{\circ}$], [90$^{\circ}$, 180$^{\circ}$], [180$^{\circ}$, 270$^{\circ}$], and [270$^{\circ}$, 360$^{\circ}$], corresponding to left front, right front, right back, and left back, with interest weights of [0.9, 0.9, 0.1, 0.1], respectively. The default DWLoss weight factor $\sigma$ is 1.0 and the default communication budget (defined in Eq. \ref{budget}) is 0.2. The setup for our experiments includes 2 Intel(R) Xeon(R) Silver 4410Y CPUs (2.0GHz), 4 NVIDIA RTX A5000 GPUs, and 512GB DDR4 RAM.

\textbf{Evaluation metrics.}
For 3D detection tasks, the intersection over union (IoU) is a common evaluation metric, calculated as the area of intersection divided by the area of union. However, IoU assesses omnidirectional perception performance. To specifically evaluate our proposed directed perception performance, we additionally introduce a metric named partial-direction intersection over union (PD-IoU). This involves dividing the BEV map into $N_{dir}$ subsets based on predefined directions, with PD-IoU separately measuring IoU within these individual subsets.

\subsection{Quantitative results}

\begin{figure*}[t]
    \centering
    \includegraphics[width=0.96\linewidth]{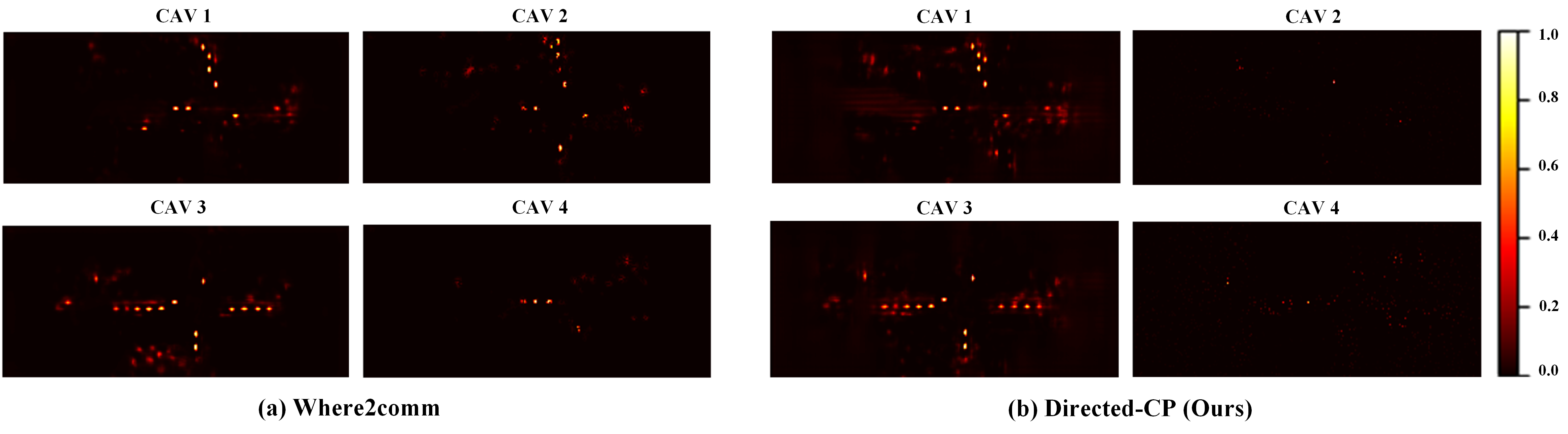}
    \caption{\textbf{Attention weight visualization} on neighboring CAVs. Where2comm distributes attention equally for omnidirectional perception, while Directed-CP focuses on features relevant to ego's interested directions.}
    \label{fig: visualization2}
\end{figure*}

\textbf{Evaluation of Directed-CP.}  We evaluate Directed-CP against baselines in the overall CP performance (AP@IoU=0.5/0.7) and in specific directions (AP@PD-IoU=0.5/0.7, interested directions are denoted with *). As shown in Table \ref{tab:quantitative_results_2}, Directed-CP uses direction-aware selective attention to reallocate communication resources, slightly outperforming the state-of-the-art Where2comm in terms of overall AP@IoU. For PD-IoU, Where2comm optimizes CP omnidirectionally, showing similar AP@PD-IoU across all directions, while Directed-CP focuses on preferred directions, achieving 18.2\% higher AP@PD-IoU=0.5 and 19.8\% higher AP@PD-IoU=0.7 than Where2comm in these directions. These results demonstrate that Directed-CP enables an ego CAV to flexibly adjust view focus and improve CP performance in the desired directions.

\begin{table*}[t]
    \large
    \caption{Quantitative results of collaborative 3D detection (communication budget = 0.2). * indicates interested directions.}
    \label{tab:quantitative_results_2}
    \renewcommand{\arraystretch}{1.1}
    \setlength{\tabcolsep}{3pt}
    \resizebox{1\linewidth}{!}{
    \begin{tabular}{l|cccc|c|cccc|c}
        \hline
        \multirow{2}{*}{\textbf{Method}} & \multicolumn{4}{c|}{\textbf{AP@PD-IoU=0.5}} & \multirow{2}{*}{\textbf{AP@IoU=0.5}} & \multicolumn{4}{c|}{\textbf{AP@PD-IoU=0.7}} & \multirow{2}{*}{\textbf{AP@IoU=0.7}} \\
        & [0°,90°]* & [90°,180°]* & [180°,270°] & [270°,360°] & & [0°,90°]* & [90°,180°]* & [180°,270°] & [270°,360°] & \\
        \hline\hline
        Lower-bound & 40.97 & 53.89 & 28.83 & 37.57 & 55.01 & 31.14 & 38.97 & 20.13 & 28.24 & 41.91 \\
        When2comm~\cite{Liu_2020_CVPR} & 34.97 & 33.56 & 19.36 & 49.96 & 53.56 & 24.81 & 24.59 & 8.75 & 40.42 & 38.70 \\
        V2VNet~\cite{10.1007/978-3-030-58536-5_36} & 57.49 & 53.62 & 28.01 & 60.36 & 67.35 & 42.61 & 41.05 & 19.54 & 36.59 & 48.22 \\
        Where2comm~\cite{hu2022wherecomm} & 51.29 & 59.38 & 48.83 & 56.27 & 79.59 & 45.86 & 44.52 & 37.89 & 48.83 & 64.96 \\
        \rowcolor{gray!5}
        \textbf{Directed-CP (Ours)} & \textbf{65.84}\textsuperscript{\color{ForestGreen}{+28.4\%}} & \textbf{65.48}\textsuperscript{\color{ForestGreen}{+10.3\%}} & 37.21 & 60.55 & \textbf{81.17}\textsuperscript{\color{ForestGreen}{+2.0\%}} & \textbf{55.76}\textsuperscript{\color{ForestGreen}{+21.6\%}} & \textbf{53.20}\textsuperscript{\color{ForestGreen}{+19.5\%}} & 28.62 & 49.98 & \textbf{66.57}\textsuperscript{\color{ForestGreen}{+2.5\%}} \\
        \hline
    \end{tabular}
    }
\end{table*}

\begin{table*}[t]
    \large
    \caption{Ablation studies on the effect of DWLoss weight factor $\sigma$ (communication budget = 0.2). * indicates interested directions.}
    \label{tab:quantitative_results_3}
    \renewcommand{\arraystretch}{1.0}
    \setlength{\tabcolsep}{2.5pt}
    \resizebox{1\linewidth}{!}{
    \begin{tabular}{l|cccc|c|cccc|c}
        \hline
        \multirow{2}{*}{\textbf{Directed-CP}} & \multicolumn{4}{c|}{\textbf{AP@PD-IoU=0.5}} & \multirow{2}{*}{\textbf{AP@IoU=0.5}} & \multicolumn{4}{c|}{\textbf{AP@PD-IoU=0.7}} & \multirow{2}{*}{\textbf{AP@IoU=0.7}} \\
        & [0°,90°]* & [90°,180°]* & [180°,270°] & [270°,360°] & & [0°,90°]* & [90°,180°]* & [180°,270°] & [270°,360°] & \\
        \hline\hline
        $\sigma=0$ & 38.28 & 51.37 & 30.71 & 27.12 & 61.63 & 14.84 & 29.16 & 14.75 & 3.38 & 31.41 \\
        $\sigma=0.5$ & 59.83 & 59.12 & 36.66 & 58.30 & 76.19 & 41.03 & 44.85 & 25.72 & 46.24 & 58.18 \\
        \rowcolor{gray!5}
        $\sigma=1.0$ & \textbf{65.84} & \textbf{65.48} & 37.21 & \textbf{60.55} & \textbf{81.17} & \textbf{55.76} & \textbf{53.20} & 28.62 & \textbf{49.98} & \textbf{66.57} \\
        $\sigma=1.5$ & 52.86 & 62.98 & \textbf{41.49} & 55.78 & 73.94 & 44.81 & 51.90 & \textbf{34.93} & 47.84 & 62.12 \\
        $\sigma=2.0$ & 49.24 & 62.46 & 32.23 & 55.14 & 73.21 & 36.97 & 48.21 & 25.51 & 42.90 & 57.18 \\
        \hline
    \end{tabular}
    }
\end{table*}

\textbf{Communication efficiency.}  We investigate how varying communication budgets affects CP performance as shown in Fig. \ref{fig: experiment} with budgets ranging from 0.01 to 0.25. Notably, below a budget of 0.1, both Directed-CP and Where2comm experience a significant drop in AP@IoU=0.7 and AP@PD-IoU=0.7 for interested directions [0, 180$^{\circ}$]. Despite this, Directed-CP slightly outperforms Where2comm overall and significantly improves perception in interested directions. At a further reduced budget of 0.01, both methods perform equally, suffering major perception degradation likely due to ultra-sparse feature maps impeding model convergence. Overall, these results highlight Directed-CP's efficiency under constrained communication resources.

\textbf{Abalation studies.} To investigate the influence of the weight factor $\sigma$ on the performance of Directed-CP, we conduct an ablation study, varying $\sigma$ from 0 to 2.0. When $\sigma$ is below 1.0, we observe a reduction in collaborative detection accuracy, particularly in less critical directions. Notably, AP@PD-IoU=0.7 for the sector [270$^{\circ}$, 360$^{\circ}$] declines to 0.03, markedly deteriorating below the lower-bound threshold. Conversely, when $\sigma$ exceeds 1.5, there is a discernible decrease in detection accuracy for both the areas of interest and the overall system. Based on these observations, a good range for $\sigma$ is between 1.0 and 1.5, which balances directed perception performance with satisfactory overall accuracy.

\subsection{Qualitative results}

\textbf{Visualization of collaborative 3D detection results.}  As shown in Fig. \ref{fig: visualization}, we display Directed-CP's collaborative detection results alongside baselines on the V2X-Sim 2.0 dataset. While Where2comm substantially improves global perception over the lower-bound, it underperforms in certain local directions, occasionally not exceeding single-vehicle outcomes, likely due to limited communication budgets and scattered focus. Conversely, our Directed-CP effectively redirects attention from less critical to key areas, significantly boosting local directional perception.

\textbf{Visualization of ego CAVs' attention weights.} As depicted in Fig. \ref{fig: visualization2}, we further compare the attention weights that the ego CAV assigned to neighboring CAVs' feature maps $W_{i,j}^{DSA}$ (defined in Eq. \ref{weights}) in two methods. With limited communication budgets, both methods query sparse features. For Where2comm, the attention weights are more uniformly assigned to other CAVs to enhance 360-degree CP. In contrast, our proposed Directed-CP pays attention to features that are more crucial to the ego CAV's interested directions, informed by other CAVs' pose information and the ego CAV's directional mask, shifting great attention from CAV 2 and 4 to CAV 1 and 3 to improve directed CP.

\section{conclusion}

In this paper, we have introduced Directed-CP, a novel CP system for ego CAVs to enhance perception in patronized directions. We have developed RSU-aided direction masking by integrating RSU's traffic detection with ego CAV's interests to identify key directions. We have also designed a proactive direction-aware attention to collect sparse feature maps from surrounding CAVs under communication budgets to improve directional perception. Additionally, we have created a direction-weighted detection loss to align perception outputs with ground truth. Experiments demonstrate that Directed-CP achieves controllable and directed perception gains under constrained communication and outperforms baselines in efficiency. Looking forward, our direction-aware framework opens new possibilities for adaptive perception, and future work could explore incorporating sophisticated traffic indicators and extending to more complex urban environments under extreme weather conditions.

\bibliographystyle{IEEEtran}
{
        \footnotesize
        \bibliography{icra2}
}

\end{document}